\documentclass[letterpaper, 10 pt, conference]{ieeeconf}  

\IEEEoverridecommandlockouts                              

\usepackage{graphicx}
\usepackage{tipa}
\usepackage{xcolor}

\usepackage[caption=false, font=footnotesize]{subfig}

\usepackage{yhmath}
\usepackage{cleveref}
\usepackage{multirow} 
\usepackage{dblfloatfix}
\usepackage{soul}

\usepackage{tikz}
\usetikzlibrary{calc}

\newcommand{\comment}[1]{\textcolor{black}{#1}}

\title{\LARGE \bf
Robotic Perception-motion Synergy for Novel Rope Wrapping Tasks
}

\author{Zhaoyuan Ma and Jing Xiao$^{1}$
\thanks{$^{1}$The authors are with Department of Robotics Engineering,
        Worcester Polytechnic Institute, Worcester, MA 01609, USA
        {\tt\small \{zma3, jxiao2\}@wpi.edu}}%
}

\begin{document}

\maketitle
\thispagestyle{empty}
\pagestyle{empty}
\begin{abstract}

This paper introduces a novel and general method to address the problem of using a general-purpose robot manipulator with a parallel gripper to wrap a deformable linear object (DLO), called a rope, around a rigid object, called a rod, autonomously. \comment{Such a robotic wrapping task has broad potential applications in automotive, electromechanical industries construction manufacturing, etc., but has hardly been studied.} Our method does not require prior knowledge of the physical and geometrical properties of the objects but enables the robot to use real-time RGB-D perception to determine the wrapping state and feedback control to achieve high-quality results. As such, it provides the robot manipulator with the general capabilities to handle wrapping tasks of different rods or ropes. We tested our method on 6 combinations of 3 different ropes and 2 rods. The result shows that the wrapping quality improved and converged within 5 wraps for all test cases. 

\end{abstract}

\begin{keywords}
Perception for Grasping and Manipulation, Perception-Action Coupling, Reactive and Sensor-Based Planning, Bimanual Manipulation, Sensor-based Control
\end{keywords}

\section{Introduction}

Manipulating flexible wire, cable, rope, or other DLOs has a wide range of applications, such as catheter inserting \cite{jayender_robot-assisted_2006}, surgical suturing \cite{schulman_case_2013}, automotive \cite{jiang_robotized_2010}, aerospace \cite{shah_planning_2018}, electromechanical industries \cite{franke_robot_2009}, and so forth. The deformable property results in high-dimensional state space for modeling DLO, which makes manipulating DLO challenging. 
\begin{tikzpicture}[remember picture, overlay]
\node at ($(current page.north) + (0,-0.5in)$)[text width=19.5cm] {\Large This work has been submitted to the IEEE for possible publication.\\ Copyright may be transferred without notice, after which this version may no longer be accessible.};
\end{tikzpicture}

Existing research for handling DLOs is focused on robot motion planning for basic tasks such as tying/untying knots \cite{suzuki_-air_2021, yamakawa_motion_2010, wakamatsu_knottingunknotting_2006}, forming a given shape \cite{yan_self-supervised_2020}, contact-based cable routing \comment{ \cite{zhu_robotic_2020, huo_keypoint-based_2022}}, inserting string and rope in a hole \cite{wang_online_2015}, and winding \cite{gobert_3dwoodwind_2022, koichiro_ito_winding_2017}. While most of the research about DLO considers quasistatic manipulation, some also address dynamic manipulation \cite{yamakawa_motion_2010, koichiro_ito_winding_2017, zhang_robot_2021}. 

Solving a DLO manipulation task usually contains three key steps: perception, modeling, and motion planning. Computer vision is often used to perceive a DLO's state. Some researchers attach AR tags along a wire harness as sampling points to detect  deformation \cite{jiang_robotized_2010}. More generally, classic algorithms, such as uniform thresholding and Canny edge detector, are applied to extract cables in a controlled environment \cite{zhu_robotic_2020}. With the development of neural networks, deep learning is used to extract features of a DLO \cite{yan_self-supervised_2020,caporali_ariadne_2022}. Tactile servoing is another approach for DLO manipulation. She {\it et al.} design a gripper with GelSight for cable following and cable insertion tasks \cite{yu_she_cable_2021}. 

Common DLO modeling is done topologically or geometrically. Topology is helpful to describe the spatial relation of a DLO fragment in a knotting task, such as presented in Wakamatsu's work \cite{wakamatsu_knottingunknotting_2006}. Different geometric methods are broadly used for other types of tasks, for example, thin plate splines \cite{schulman_case_2013}, parameterized curve with minimum energy-based scheme \cite{shah_planning_2018}, multi-link system \cite{yamakawa_motion_2010}, \comment{truncated Fourier series model \cite{huo_keypoint-based_2022}}. A bi-directional long short-term memory (LSTM) is also used to model the structure of a chain-like mass-spring system \cite{yan_self-supervised_2020}. Alternatively, it is possible to bypass the modeling step with deep learning to create low-level joint control directly from input sensor data, as suggested by Suzuki {\it et al.} \cite{suzuki_-air_2021}, who use a convolutional auto-encoder (CAE) and LSTM structure. The system uses RGB images and proximity sensor information as input to generate robot joint angles directly.

Once the model is ready, a task-oriented motion planning method is introduced to solve the problem, such as learning from demonstration \cite{schulman_case_2013}, model predictive path integral (MPPI) control \cite{yan_self-supervised_2020},  planning based-on  angular contact mobility index (ACMI) \cite{zhu_robotic_2020}, \comment{hierarchical action primitives \cite{huo_keypoint-based_2022}}.

\comment{There are several papers investigating DLO wrapping tasks from dynamic or quasistatic aspects. Lee {\it et al.} focused on testing the performance of simulating a high volume of possible contacts \cite{lee_parallelized_2021}. Göbert {\it et al.} \cite{gobert_3dwoodwind_2022} designed a customized end-effector to attach a winding filament to a cyclical mold with a rotation axis and planned the path of the mold. Ito {\it et al.} studied using one whipping motion to wind a whip onto a target object with dynamic manipulation \cite{koichiro_ito_winding_2017}. }

However, there is a lack of study to enable an off-the-shelf, general-purpose robot manipulator to perform general wrapping manipulation of a DLO around another object. 

In this paper, we address the open problem of enabling a general-purpose robot manipulator with a simple parallel gripper to autonomously wrap a DLO around the object based on synergizing real-time perception and robot motion planning and control without requiring prior physical and geometrical information of both the DLO and the object. We are interested in providing a general robotic wrapping capability that can be applied to DLOs of varied materials, including different kinds of ropes, flexible cables, fibers, and so on. Specifically, the paper presents a novel approach for general-purpose robot wrapping operations with the following characteristics:
\begin{itemize}
    \item It uses real-time perception of the objects and wrapping state to determine and adapt the robot motion for accomplishing and improving wrapping operations to achieve high-quality results.
    \item It re-uses and adjusts the canonical motion of making a single wrap of the DLO around the object to be flexible to the length of the DLO (thus the length of the coil as the wrapping result) and to enable constant check and improvement of wrapping quality with feedback control.
    \item Hence, it is able to achieve high-quality wrapping without requiring prior knowledge of the physical and geometrical properties of the DLO and the other object. 
\end{itemize}

This paper is organized as follows. \Cref{section:task_definition} defines the problem and setup. 
\Cref{section:approach,section:perception,section:synergy} introduces our approach in detail.
\Cref{section:exp_n_result} describes our experiments and presents the results. \Cref{section:conclusions} concludes the paper.

\section{Task description}\label{section:task_definition}

\begin{figure}[t]
    \centerline{\includegraphics[width=0.4\textwidth]{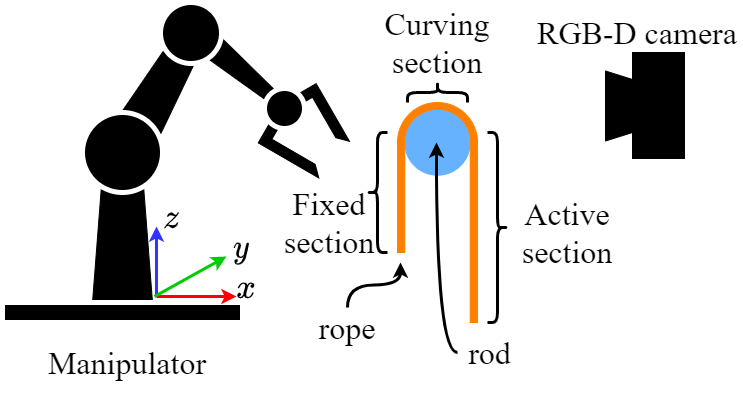}}
    \caption{The side view of the task setup. \comment{The system is composed of a manipulator, a rope over a rod, and an RGB-D camera.}}
    \label{fig:task}
\end{figure}

We define the task we study as wrapping a DLO, called the ``rope'', around a given rigid object, called the ``rod''. A coordinate system is set up at the base of the manipulator. The positive directions along the $x$, $y$, and $z$ axes are defined as ``front'', ``left'', and ``up'' respectively from the manipulator's perspective. The rod is set in front of the manipulator. The manipulator has a gripper installed as the end-effector. In the initial state, the rope rides over the rod. We divide the rope into three sections: the fixed section, the curving section, and the active section. The fixed section is attached to a fixture or held by the manipulator without moving during the wrapping. The curving section has already been wrapped around the rod. The active section has sufficient length to create a few more wraps. The system could only obtain information through an RGB-D camera set to face the rod and the manipulator. Fig.~\ref{fig:task} shows the side view of the setup. The goal of the task is for the manipulator to wrap the rope around the rod to create a helix that is tight in both radial and axial directions. Neither the dimensions nor the materials of the rod and the rope are known to the robot system. 

\begin{figure}[t]
    \centerline{\includegraphics[width=0.320\textwidth]{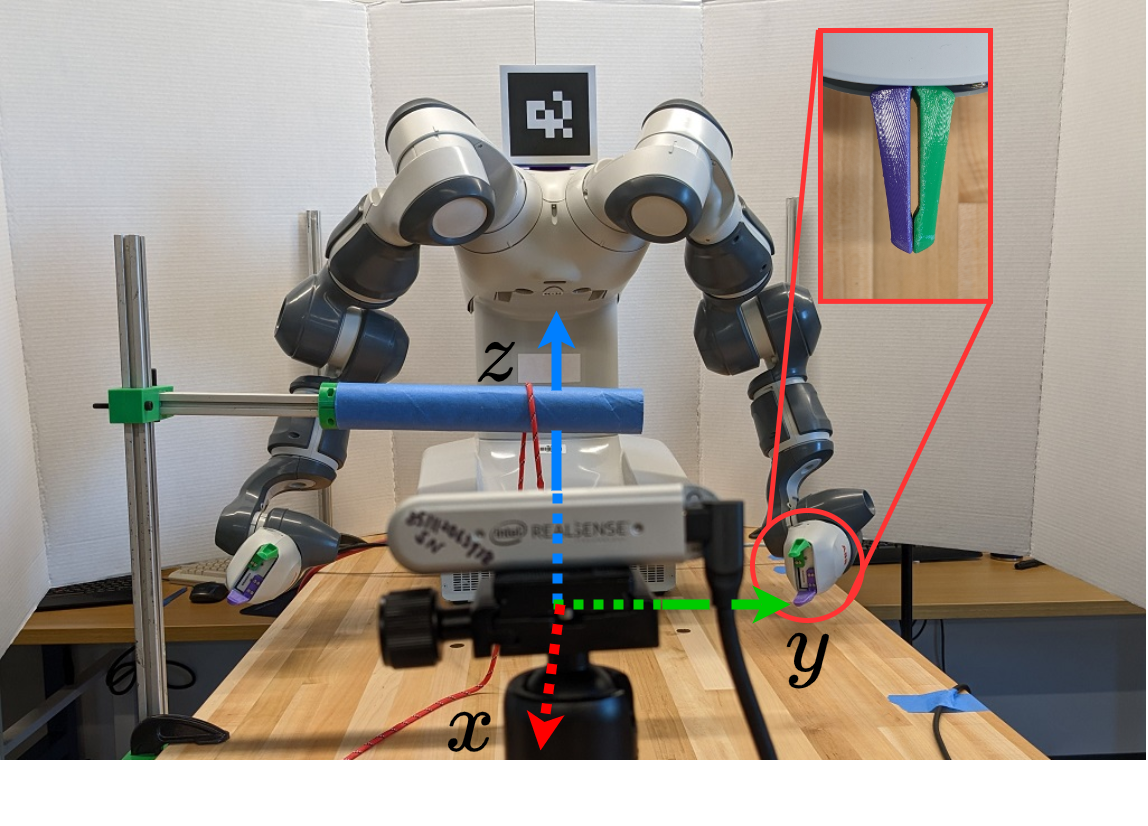}}
    \caption{The front view of the experiment setup. \comment{In order from closest to farthest are the camera, the rope (red), the rod (blue), and the YuMi manipulator. The world coordinate system is set on the tabletop.}}
    \label{fig:setup}
\end{figure}

\section{Approach}\label{section:approach}

We set up a real environment \comment{to realize the task}, as shown in Fig.~\ref{fig:setup}. A dual-arm manipulator (YuMi IRB 14000, ABB) is placed on a tabletop for the task. The fingers of the manipulator have been modified to have a $2.5$mm wide $30$mm long slot between two fingers when the gripper is fully closed. This allows the rope to slide in between while keeping the tension of the rope. A support structure is mounted in front of the manipulator to install the rod. An RGB-D camera (RealSense Depth Camera D415, Intel) is placed to face the manipulator. The manipulator and the camera are connected to a desktop computer as the controller.

Our system first conducts {\bf rod estimation}: using collected RGB images and point cloud data to estimate the position, orientation, and dimensions of the rod with respect to the manipulator automatically. This step is detailed in \Cref{section:rod_estimation}.

Next, our system processes {\bf rope estimation} by using RGB images and the rod estimation result to obtain the rope's color and diameter information. This step is described in \Cref{section:rope_estimation}.

Subsequently, our system conducts a single wrap of the rope around the rod. \comment{It involves the following major perception and motion procedures:} 
\begin{enumerate}
    \item {\bf Grasp point selection: }using RGB images to search for a grasp point along the rope. 
    \item {\bf Motion adjustment for wrapping: }generating the manipulator's wrapping motion path based on adjustable parameters.
    \item {\bf Auxiliary motion generation to facilitate wrapping: }generating picking and releasing motions for the manipulator to perform before and after the wrapping motion respectively to complete the whole process.
    \item {\bf Real-time perception and feedback control: }\comment{using RGB images to estimate the wrapping result of the motion and adjust the parameters of the motion path generator in procedure 2). }
\end{enumerate}
\noindent Those procedures are described in \Cref{section:gp_selection,section:wrapping,section:aux,section:fb_control} in detail. Our system repeats the above \comment{four-step perception-motion synergy} to wrap the rope around the rod while improving the wrapping quality until it can produce a tight helix wrap satisfactorily.

\section{\comment{Rod and rope estimation}}\label{section:perception}

\comment{In this section, we present our approach for estimating the rod and the rope as a preprocessing step for a wrapping task.}

\subsection{Rod estimation}\label{section:rod_estimation}

\begin{figure}
     \centering
     \subfloat[The workspace is captured by the RGB-D camera to generate colorized point cloud.\label{fig:raw_pcd}]
        {\includegraphics[width=0.45\linewidth]{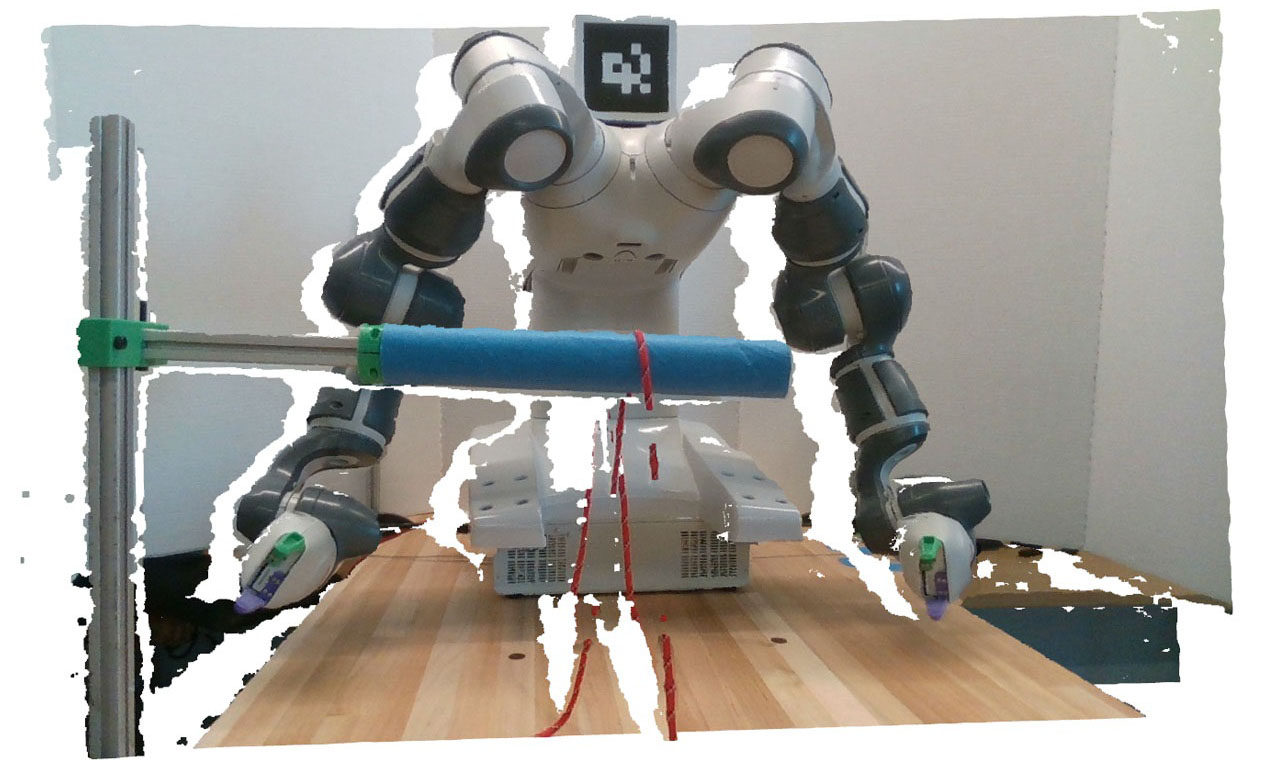} }
     \hfill
     \subfloat[The robot and the background are subtracted. The data points are clustered by distance.\label{fig:dbscan}]
        {\includegraphics[width=0.45\linewidth]{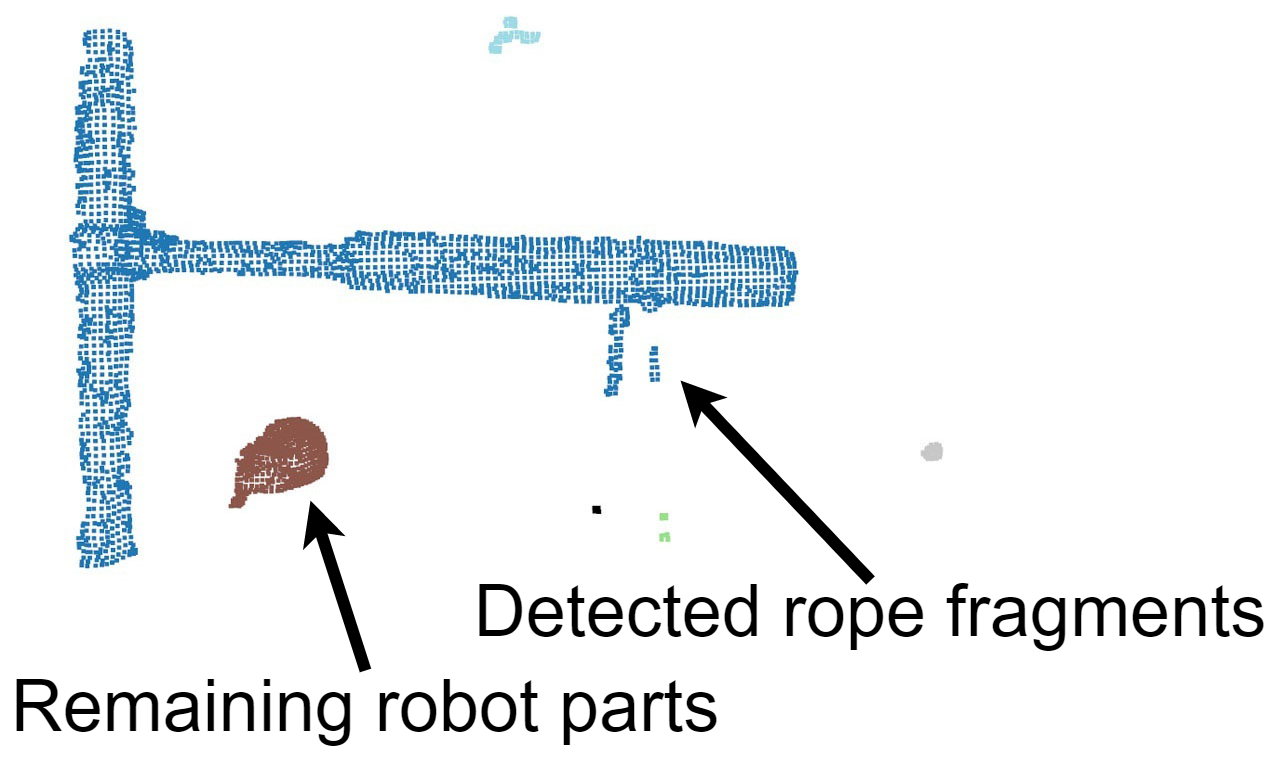} }

    \vspace{-1\baselineskip}

     \subfloat[The cluster with the closest distance toward the camera is selected.\label{fig:rod_n_support_pcd}]
          {\includegraphics[width=0.45\linewidth]{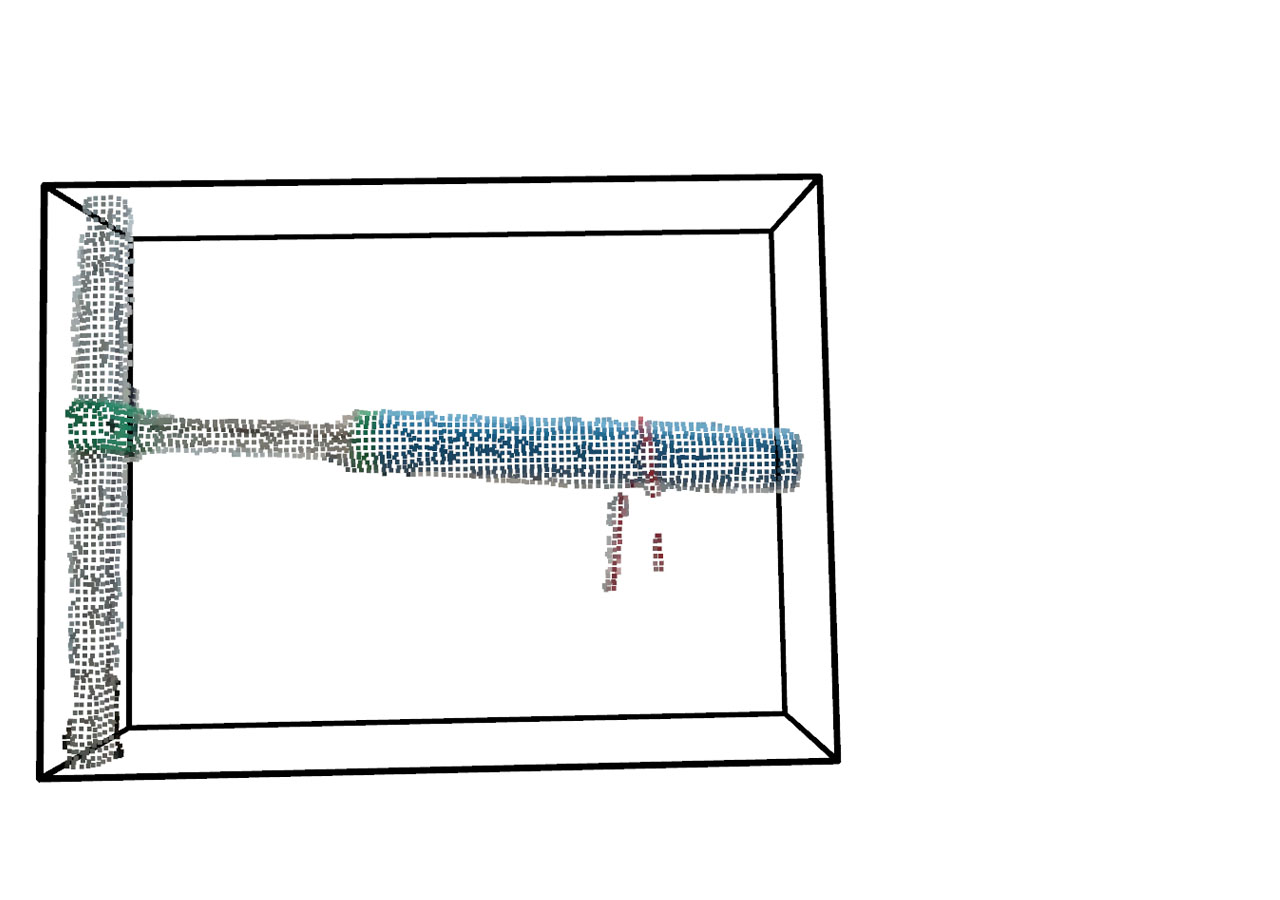} }
     \hfill
     \subfloat[An image mask is generated from the selected cluster. \label{fig:rod_n_support_mask}]
         {\includegraphics[width=0.45\linewidth]{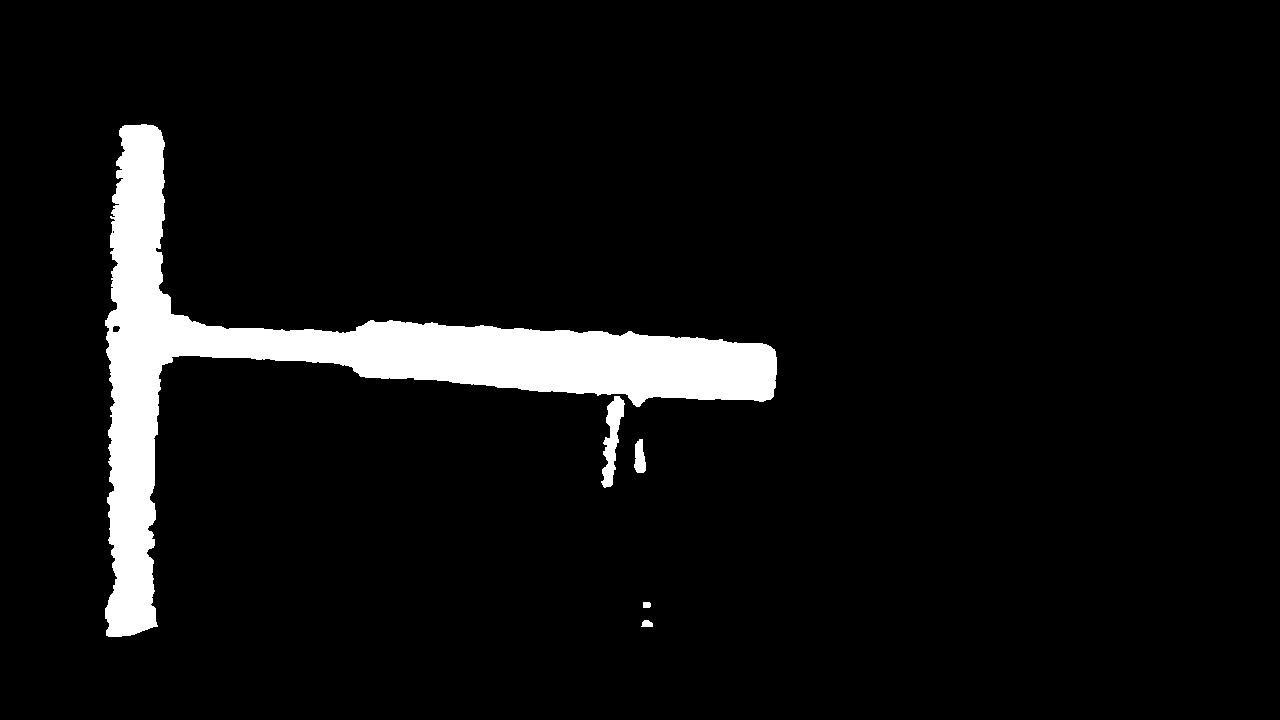} }

     \vspace{-1\baselineskip}

     \subfloat[The maximum inscribed rectangle (red) is added to estimate the rod on the 2D image plane.\label{fig:inscribed_rectangle}]
         {\includegraphics[width=0.45\linewidth]{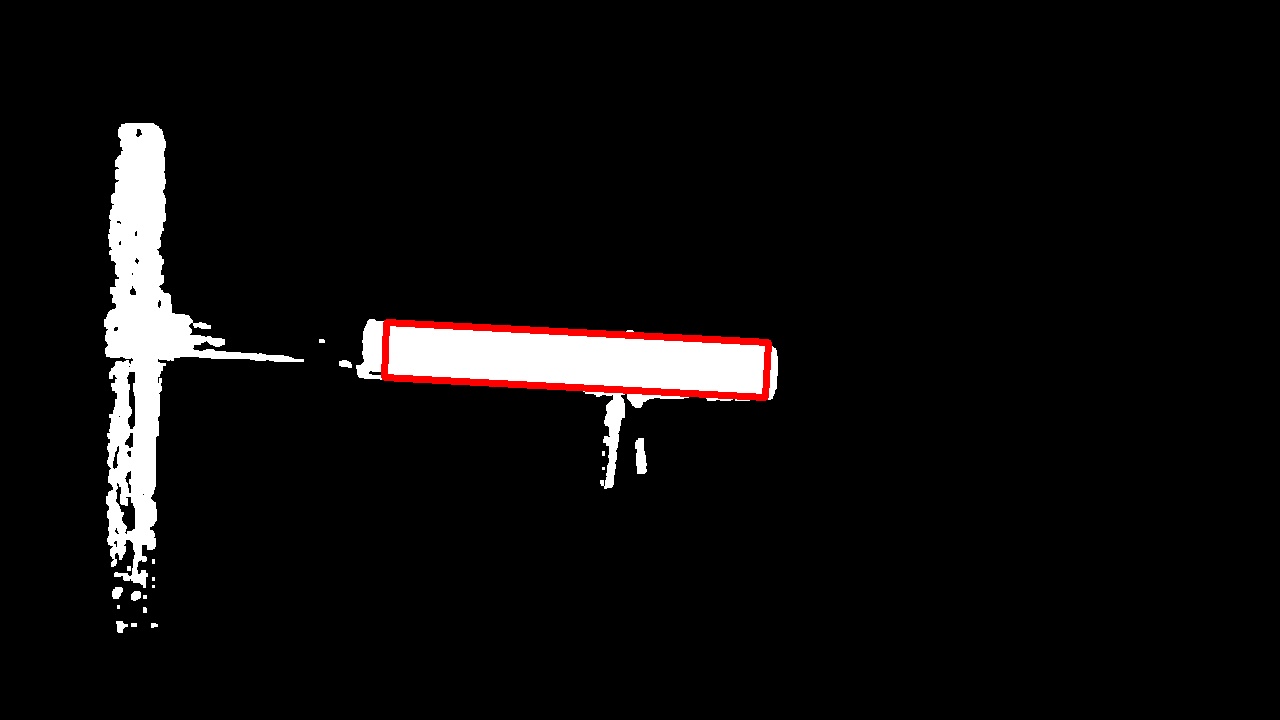} }
     \hfill
     \subfloat[The half-cylinder template (yellow) is matched to the point cloud by applying ICP. \label{fig:icp_result}]
        {\includegraphics[width=0.45\linewidth]{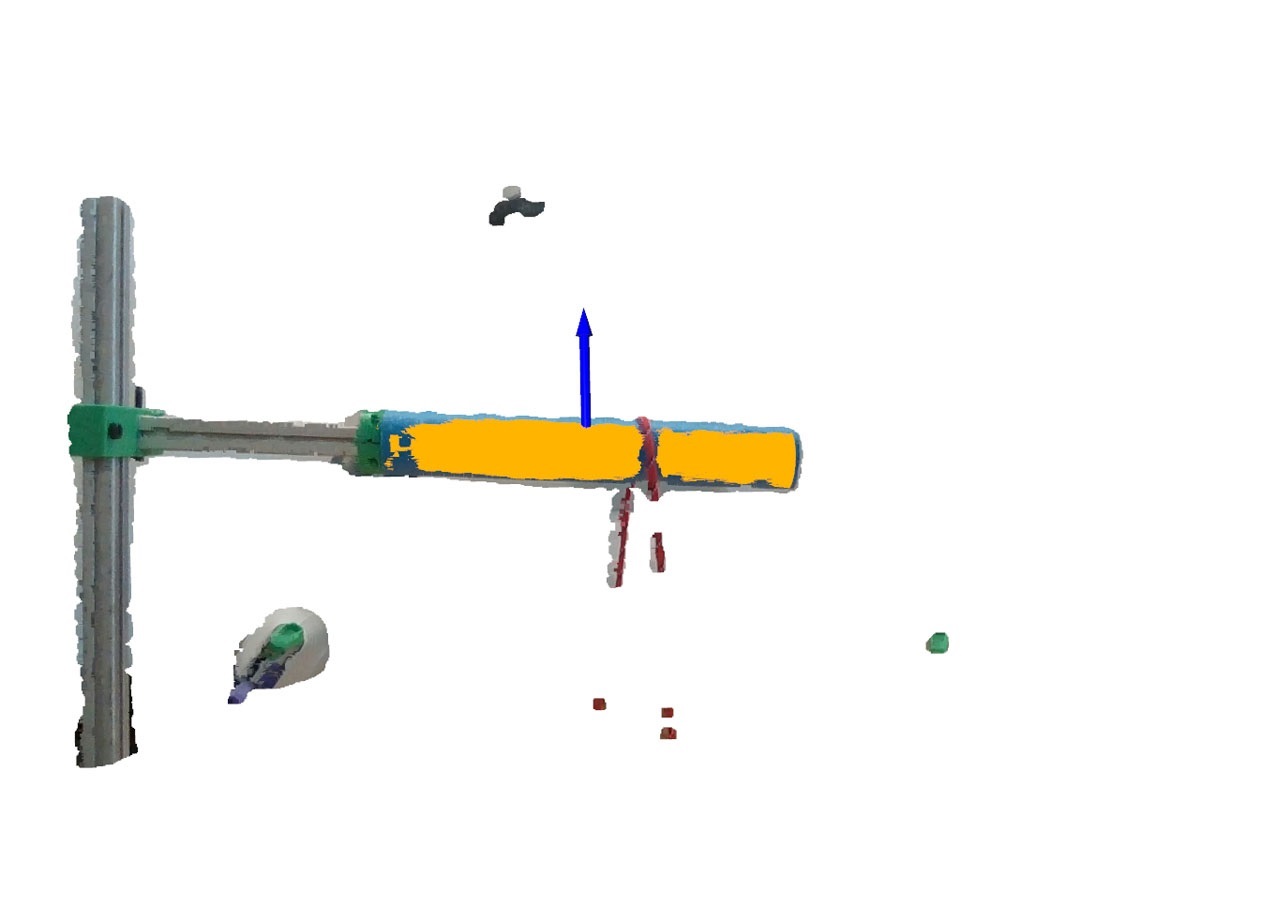} }
     
        \caption{Key steps to estimate the rod's dimension and pose.}
        \label{fig:find_rod}
\end{figure}

\comment{We developed a process to identify the rod's dimensions, position, and orientation. Note that the rod's axis does not need to be parallel to the $yz$-plane (Fig.~\ref{fig:setup}).} The process starts with localizing the camera with the fiducial marker \cite{ar_track_alvar} that was attached to the robot. The RGB\nobreakdash-D camera provides a colorized depth map of the workspace (Fig.~\ref{fig:raw_pcd}). Let $P$ denote the set of all the 3D points in the map, and let $Q$ be the set of all 2D pixels with color information in the map. Each data point within the frame can be represented as $(p_x, p_y, p_z, q_x, q_y, c)$, where $(p_x, p_y, p_z)\in P$ is the position in the camera coordinate system, $(q_x, q_y, c)\in Q$ contains the pixel location on the image plane and the color $c$. \comment{For any given $(p_x,p_y,p_z)$, there is a unique $(q_x, q_y)$ corresponding to it, and vice versa.}

The system extracts the points between the robot and the camera, and above the tabletop from the point cloud data. The system downsamples the extracted point cloud \comment{to further reduce the number of data points and then applies DBSCAN \cite{ester_density-based_1996} to create clusters according to the points' position. An example result is shown as Fig.~\ref{fig:dbscan}. Knowing the rod is the closest object to the camera, the system keeps the cluster with the shortest distance to remove the remaining robot parts.} A 3D bounding box is created around this cluster (Fig.~\ref{fig:rod_n_support_pcd}). All points within the 3D bounding box are collected. White pixels in Fig.~\ref{fig:rod_n_support_mask} indicate the selected data points on the 2D image plane. \comment{The system uses K-mean \cite{macqueen_methods_1967} to classify the data points into 2 clusters according to their hue, assuming that the rod has a more uniform color and occupies a larger portion of the image. Subsequently, the system retains the larger cluster, which contains the data points belonging to the rod.} A maximum inscribed rectangle is used to fit the kept points, as shown in Fig.~\ref{fig:inscribed_rectangle}.

Finally, the points within the rectangle are treated as points on the rod, \comment{with ${Q'\subset Q}$ denoting the pixels and ${P'\subset P}$ being the corresponding 3D points.} $P'$ is used to estimate the radius $r_{rod}$ and the length $l_{rod}$ of the rod. Our system creates a half-cylinder surface template based on the estimation and employs ICP \cite{besl_method_1992} to match it to $P'$ (Fig.~\ref{fig:icp_result}). At this point, the estimates of the rod's center position $(x_{rod},y_{rod},z_{rod})$, the rod's orientation, and $r_{rod}$ are obtained.

\subsection{Rope estimation}\label{section:rope_estimation}
\begin{figure}
\begin{minipage}{.25\textwidth}
  \subfloat[Pixels on the rod are selected from the image (red rectangle). \label{fig:img_w_box}]
    {\includegraphics[width=.95\linewidth]{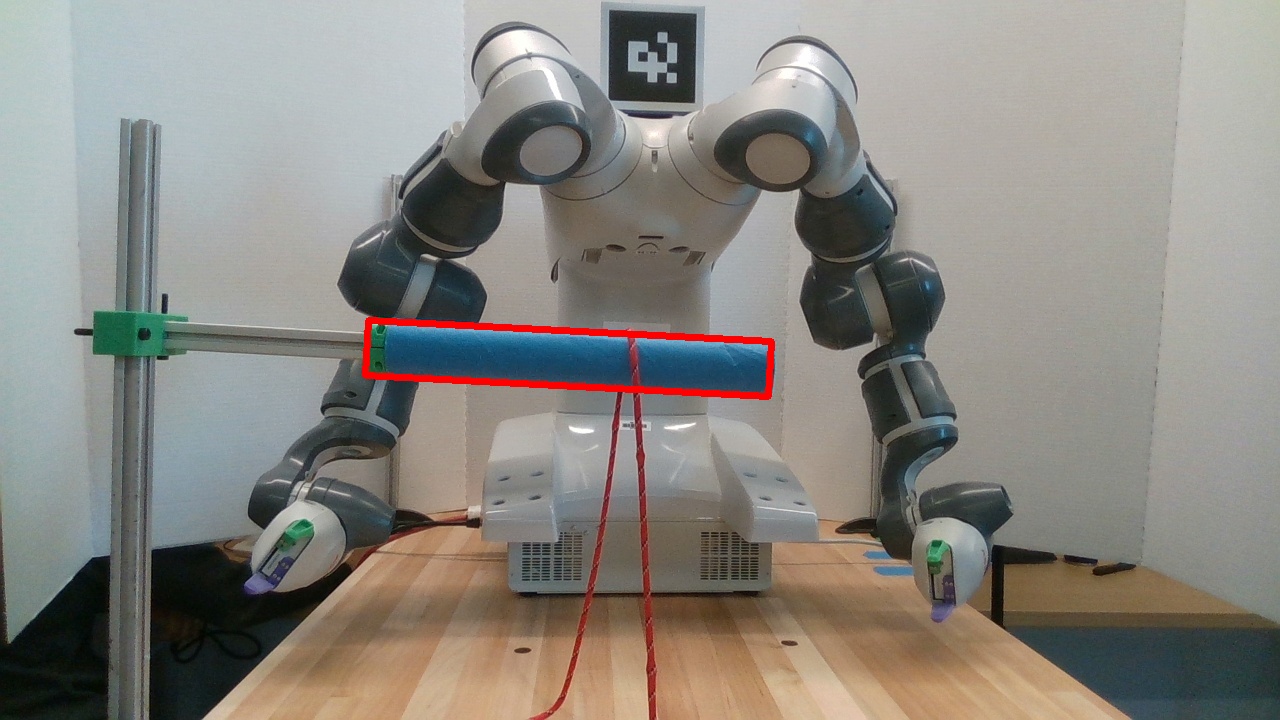} }

\end{minipage}
\hspace{.005\textwidth}
\begin{minipage}{.22\textwidth}
  \subfloat[The rope on the rod is obtained via thresholding with its hue feature. \label{fig:rope_mask}]
    {\includegraphics[width=.95\linewidth]{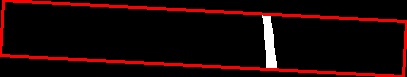} }

  \subfloat[A contour is created to represent the piece of the rope. \label{fig:rope_contour}]
    {\includegraphics[width=.95\linewidth]{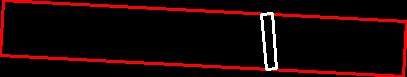} }

\end{minipage}
\caption{Estimating the rope's width and color.}
\label{fig:rope_width}
\end{figure}

\comment{The point cloud only captures small fragments of the rope due to its limited 3D resolution (see Fig.~\ref{fig:dbscan}). Therefore, the system uses RGB images for rope-related processes.} Our system estimates the color (hue range) and the diameter $d$ of the rope using $Q'$ (highlighted by the red rectangle in Fig.~\ref{fig:img_w_box}). It extracts the hue channel of $Q'$ to create a histogram. Otsu's method \cite{otsu_threshold_1979} is applied to the histogram to find a threshold that can separate $Q'$ into the rope and the rod. Then a Gaussian function $N(\mu,\sigma)$ is used to approximate the normalized rope's hue histogram. The hue range of the rope is chosen as $[\mu-3\sigma,\mu+3\sigma]$.

The hue threshold found above is also applied to $Q'$ to create a binary mask of the rope. The result is shown in Fig.~\ref{fig:rope_mask}. A minimum area rectangle is generated to enclose the selected area, as in Fig.~\ref{fig:rope_contour}. The \comment{width (short edge)} of the rectangle is taken as $d$.

\section{\comment{Perception-motion synergy for wrapping}}\label{section:synergy}

In this section, we introduce the perception and motion procedures that iterate to generate and improve wraps.

\subsection{Grasp point selection}\label{section:gp_selection}

\begin{figure}[btp]
     \centering
     \subfloat[Selected region that contains the rope.\label{fig:rope_region}]
         {\includegraphics[width=0.45\linewidth]{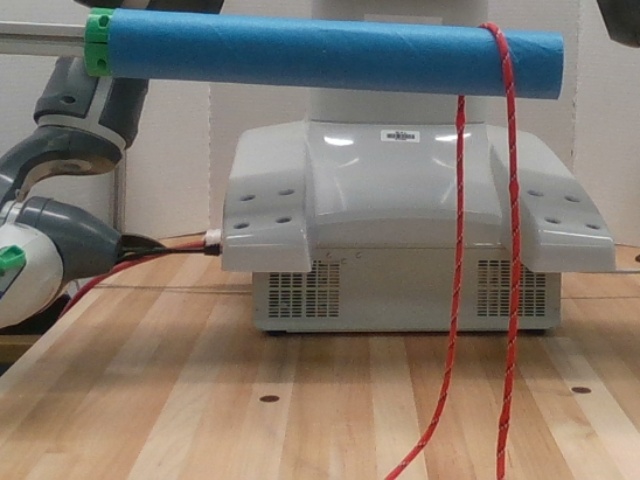} }
    \hfill
     \subfloat[Mask $M_1$ generated by using Ariadne+. \comment{The red arrow points to the detection defect.} \label{fig:ariadne}]
         {\includegraphics[width=0.45\linewidth]{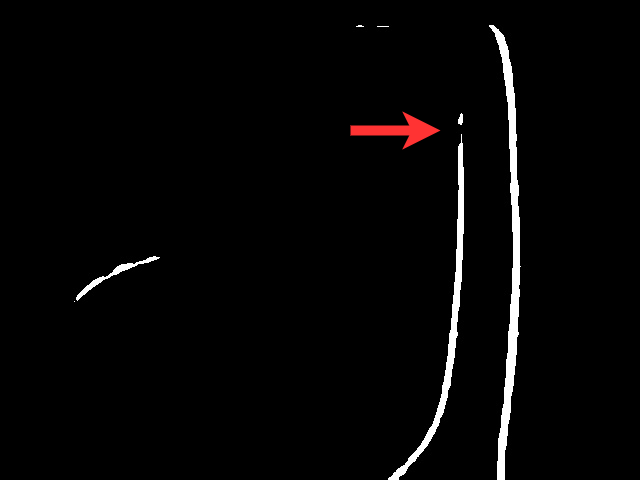} }
         
     \subfloat[Mask $M_2$ generated by using the rope's hue feature. \label{fig:hue_mask}]
         {\includegraphics[width=0.45\linewidth]{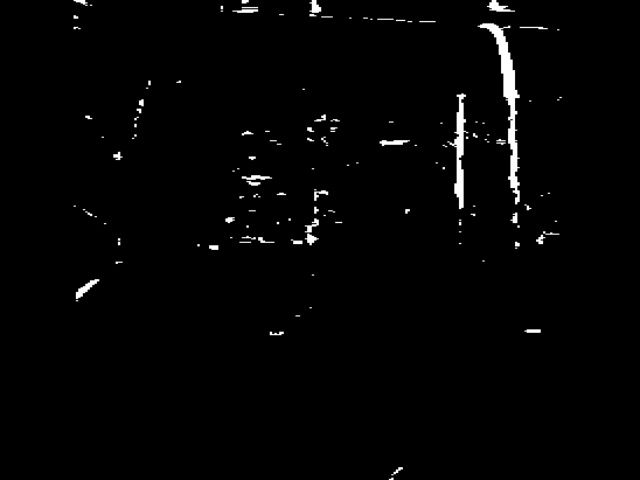} }
    \hfill
     \subfloat[Masks are combined and skeletonized. \label{fig:rope_skeleton}]
         {\includegraphics[width=0.45\linewidth]{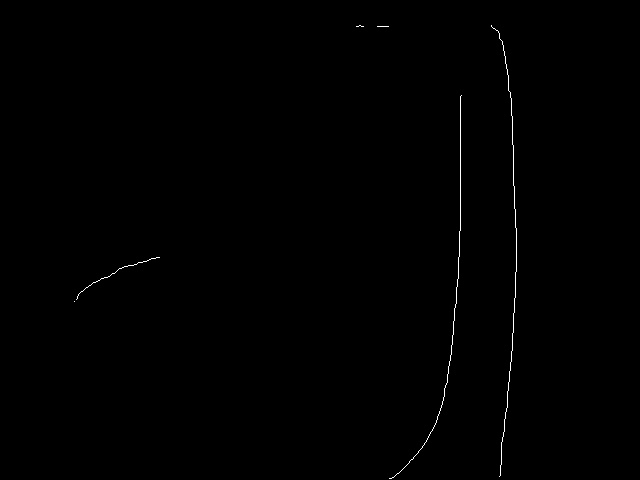} }

     \subfloat[The fixed section (green), the active section (blue), and the grasp point (red) on the fixed section are found on the 2D image plane. \label{fig:found_ropes}]
         {\includegraphics[width=0.45\linewidth]{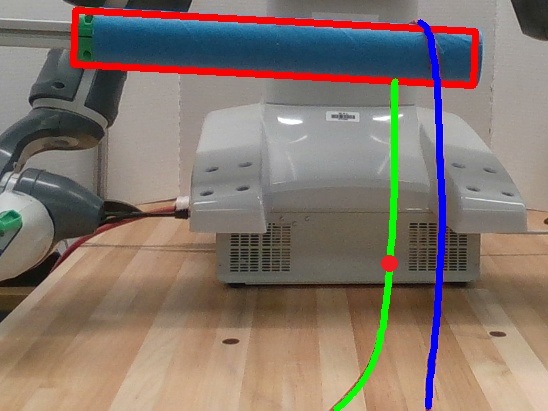} }
     \hfill
     \subfloat[The 3D position of the grasp point is located for the follow-up robot motion planning. \label{fig:grasping_result}]
         {\includegraphics[width=0.45\linewidth]{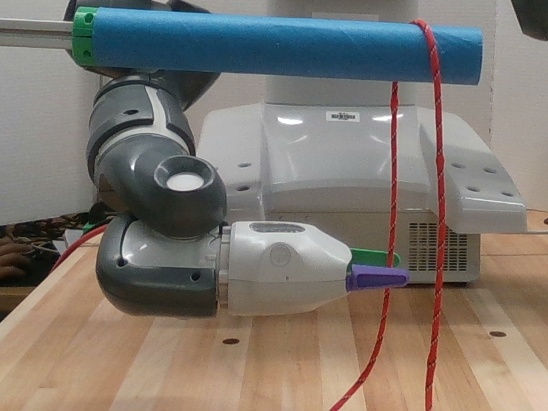} }
         
 \caption{Example of grasp point selection for the fixed section.}
 \label{fig:grasping_point_selection}
\end{figure}

Wrapping a rope typically requires grasping both the fixed section and the active section and moving the active one around the rod. Finding the grasp points on the two sections is performed with an unwrapped rope. For each additional wrapping motion, only the grasp point on the active section needs to be updated. \comment{This step needs an input $l_{gp}$ (in millimeters). It measures from the grasp point up to the lower edge of the rod, along the designated rope section. For the fixed section, $l_{gp}$ is set as a constant. For the active section, the $l_{gp}$ selection is given in \Cref{section:wrapping}}.

Grasp point selection starts with extracting the two sections of the rope from the image. A sub-image of Fig.~\ref{fig:img_w_box} is created as shown in Fig.~\ref{fig:rope_region}, by extending the bounding box of the rod at both sides and downward. The system employs Ariadne+'s \cite{caporali_ariadne_2022} pre-trained DeepLabV3+ \cite{chen_encoder-decoder_2018} to create a binary mask $M_1$ from the sub-image. Compared to traditional computer vision methods, this deep learning approach suggests possible rope areas with less noise. However, we observed that extraction defects may happen. For example, Fig.~\ref{fig:ariadne} shows an incomplete detection on the fixed section that is near the rod. Therefore, our system applies the rope's hue range as the threshold to the sub-image to create another binary mask $M_2$, as shown in Fig.~\ref{fig:hue_mask}. $M_2$ is used to connect detected rope segments in $M_1$ as much as possible. \comment{For a white pixel $(q_x,q_y)$ in $M_1$, if $M_2$ has a set of vertically connected white pixels that across at the same position $(q_x,q_y)$, all the pixels in this set are added to $M_1$.} The updated $M_1$ is then skeletonized \cite{zhang_fast_1984} to reduce detected objects to 1-pixel width, as shown in Fig.~\ref{fig:rope_skeleton}. Finally, our system searches for all lines from the bottom of the skeletonized mask, extracts the two longest lines, and calculates their center. The one near the right gripper is taken as the fixed section (highlighted with green), and the one near the left gripper is taken as the active section (highlighted with blue), see Fig.~\ref{fig:found_ropes}.

After the two sections of the rope are detected, our system converts $l_{gp}$ to the measurement in pixels and finds the corresponding point along the designated rope section. \comment{In order to find the corresponding point on the rope, our system assumes that (i) the tangent point of a section is on the rod, and its position can be obtained by using the rod's estimation result, and (ii) for two points on a section, they have the same $x$ value, and the distance on the 2D image and the distance in 3D are uniformly scaled. The assumptions hold because the two rope sections are kept vertical by an extra motion of the manipulator, as described in \Cref{section:aux}, in addition to gravity. Hence, both fixed and active sections within the sub-image are roughly parallel to the $yz$-plane and tangent to the rod, as in Fig.~\ref{fig:task}.} In Fig.~\ref{fig:found_ropes}, the found point in the image plane is represented by a red dot. The position information of the pixel is converted back to the world coordinate system and used to move the robot's gripper, as shown in Fig.~\ref{fig:grasping_result}.

\subsection{Motion adjustment for wrapping}\label{section:wrapping}

The wrapping motion is the most crucial as it determines the quality of the generated wrapping helix. It is performed in the presence of the estimation error or uncertainty of the rod's pose, dimension, the grasp points along the rope, and the unknown physical properties of the rope. 

Our approach is to create a spiral curve with just a few parameters for the gripper to follow and then adjust the parameter values based on perception feedback. The goal is to achieve a resulting wrapping motion that can overcome the estimation errors and achieve a high-quality wrap in spite of the unknown physical properties of the rope. 

\begin{figure}[h]
    \subfloat[The side view of the section of the rod and the unwrapped rope. \label{fig:rope_n_rod_side}]
        {\includegraphics[width=.45\linewidth]{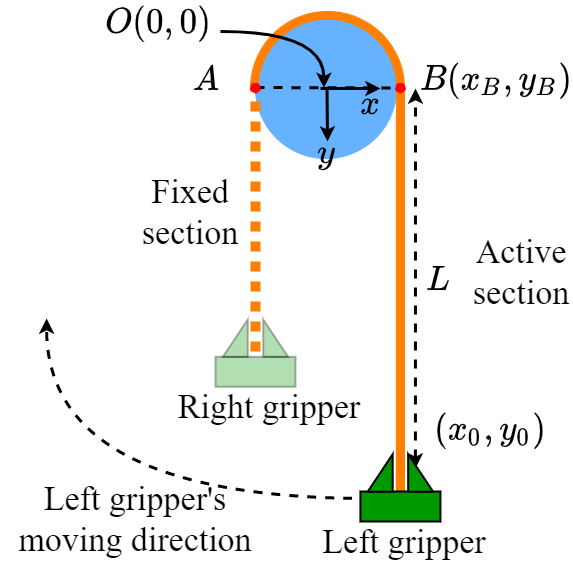} }
    \hfill
     \subfloat[The side view of the gripper's path (magenta dashed line). \label{fig:spiral_side}]
        {\includegraphics[width=.45\linewidth]{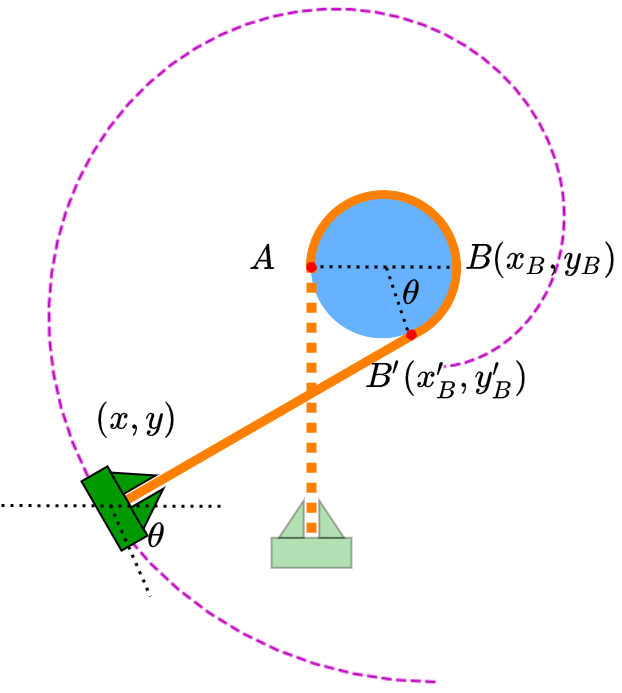} }
    
    \subfloat[The front view of the rope's diameter and the wraps' advance. The translucent orange section indicates the active section before the wrapping. The opaque one on the right is after wrapping. \label{fig:spiral_front}]
        {\makebox[0.95\linewidth]{\includegraphics[width=.45\linewidth]{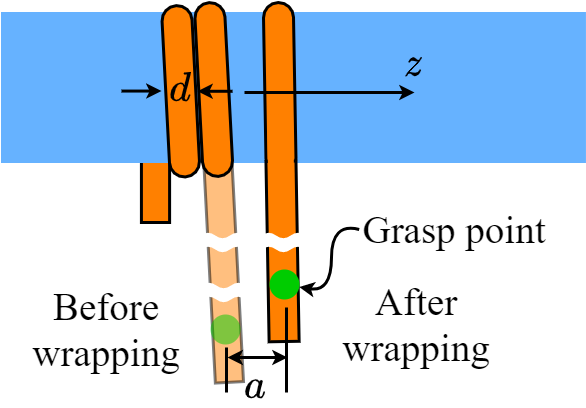} }}

\caption{Spiral curve for the robot's gripper.}
\label{fig:spiral}
\end{figure}

To design the canonical spiral curve for the gripper, we assume a rope hangs over the rod naturally due to gravity, creating contact with the rod on the upper half of the cylinder and leaving the surface of the rod tangentially at points $A$ and $B$, as shown in Fig.~\ref{fig:rope_n_rod_side}. Let $O$ be the center of the rod's cross-section where the rope lies. For the convenience of deriving the spiral function, we define a new 2D coordinate system with the origin at $O$ and the $x$ and $y$ axes as indicated in Fig.~\ref{fig:rope_n_rod_side} on the cross-section. The 3D coordinate of $O$ can be obtained from its relation to the rod's center.

The fixed section starts from point $A$, extends downwardly, and is held by the robot's right gripper. The active section is extended from point $B$ to the grasp point $(x_0,y_0)$, with the length \comment{$l_{gp}=L$, as shown in Fig.~\ref{fig:rope_n_rod_side}}. We define $R=r_{rod} + \epsilon$, where $\epsilon$ is a variable for correcting the estimation error in $r_{rod}$. \comment{$R$ is considered in its entirety. Its adjustment is discussed in \Cref{section:fb_control}.}  We define $L=2\pi R+L'$, where $L'$ is a safe distance that considers the gripper's size. Once the left gripper starts to make a wrap, the tangential contact point $B$ moves to $B'$. Note that the wrapping angle $\theta\in[0,2\pi]$ is the angle $\angle B'OB$. As an additional $\wideparen{BB'}=\theta R$ of the rod is covered by the rope, the distance between the gripper to the tangential point $B'$ is reduced to $L-\theta R$, as in Fig.~\ref{fig:spiral_side}. 

Now we consider the following spiral curve with respect to the rod coordinate system:
\begin{equation}\label{eq:spiral}
\left\{\begin{array}{l}
    x = R\cos\theta-(2\pi R+L'-\theta R)\sin\theta \\
    y = R\sin\theta+(2\pi R+L'-\theta R)\cos\theta\\
    z = a\theta/2\pi
\end{array}\right.
\end{equation}
\noindent where $a$ is the displacement of the gripper along the rod's axial direction for one wrap (Fig.~\ref{fig:spiral_front}), which can be decided through feedback (see \Cref{section:fb_control}). 
The 2D projection of this spiral path for the gripper is shown as the magenta dashed curve in Fig.~\ref{fig:spiral_side}. 
Our system searches for the safe distance $L'\in[L'_{\min},L'_{\max}]$ so that the wrapping path has a feasible inverse kinematics (IK) solution.

Note that the goal of the wrapping is to create wraps that are tight along the radial and axial directions of the rod, such that the wrap along the radial direction is as close to the radius of the rod as possible, and the distance along the axial direction from the center of two adjacent wraps equal to the diameter of the rope.

\begin{figure}[t]
     \centering
     
     \subfloat[The gripper wraps with a fixed orientation, causing the rope to tangle the fingers. \\ \label{fig:rope_entangled}]
         {\includegraphics[width=0.45\linewidth]{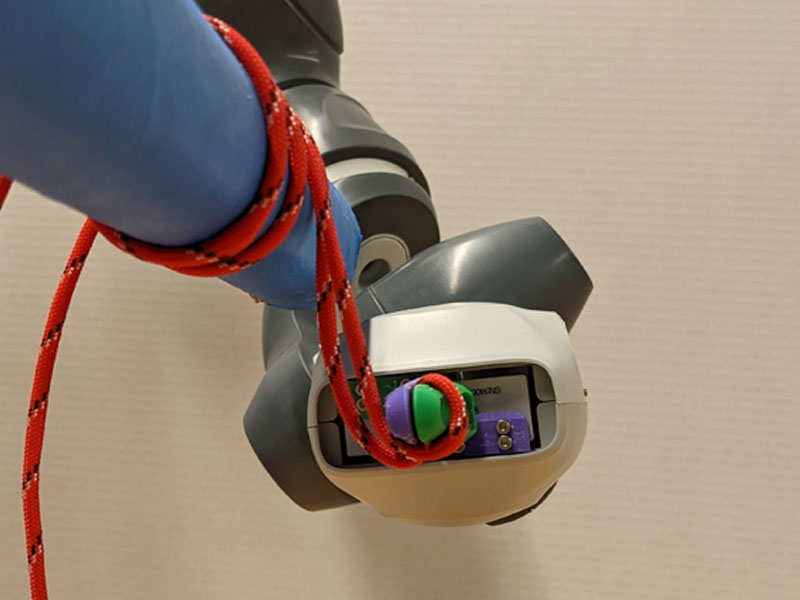} }
     \hfill
     \subfloat[The gripper wraps with the orientation adapted to the spiral curve. \label{fig:gripper_orientation}]
         {\includegraphics[width=0.45\linewidth]{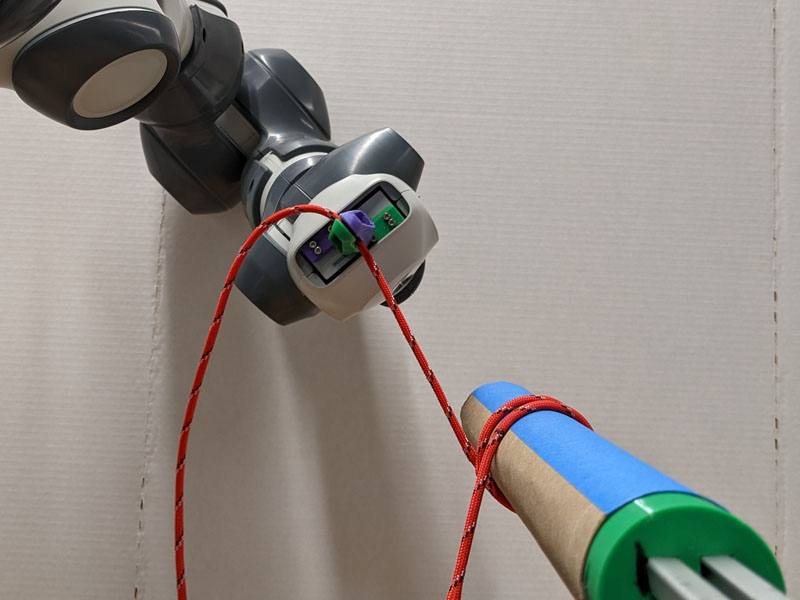} }

     \vspace{-1\baselineskip}

     \subfloat[The gripper finishes the wrapping motion and leaves a part of the active section hanging over the rod. \label{fig:rope_flip}]
         {\includegraphics[width=0.45\linewidth]{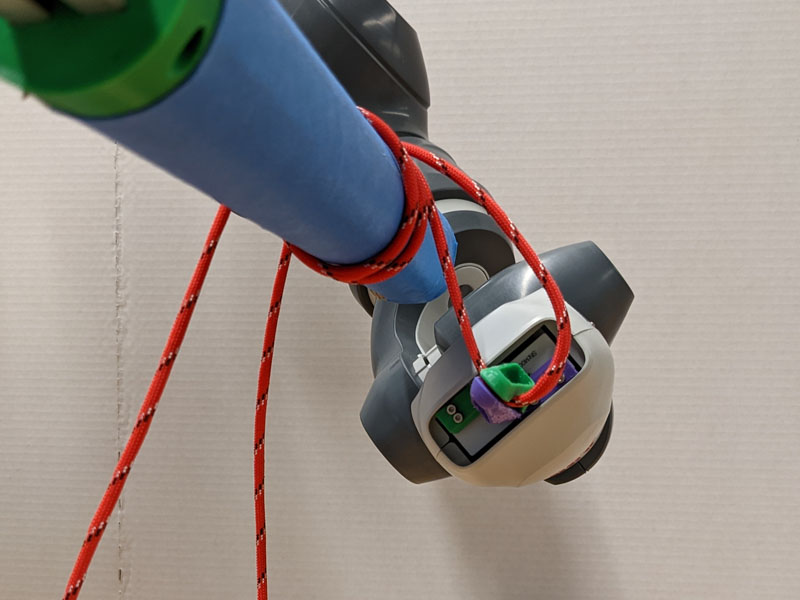} }
     \hfill
     \subfloat[The gripper follows a parameterized spiral curve to create a wrap and straight-line path to straighten the rope. \label{fig:rviz}]
         {\includegraphics[width=0.45\linewidth]{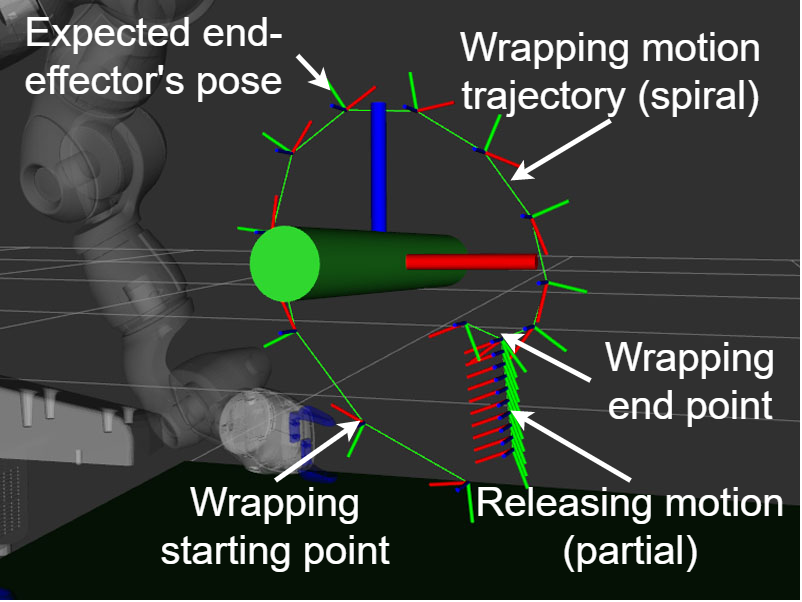} }
    \caption{The wrapping motion.}
    \label{fig:wrapping_motion}
\end{figure}

 During the wrapping process, when the wrapping angle changes to $\theta$, the gripper rotates to the same angle $\theta$ about the rod's axis simultaneously, as indicated by its orientation in Fig.~\ref{fig:spiral_side}. Without this rotation, the rope tends to entangle the fingers and hinder the gripper's opening motion at the end of the wrap, as shown in Fig. \ref{fig:rope_entangled}. The results of wrapping with the change of the gripper orientation are shown in Fig.~\ref{fig:gripper_orientation} ($\theta=120^\circ$) and Fig.~\ref{fig:rope_flip} ($\theta=330^\circ$). At this stage, the position and orientation of the robot gripper along the spiral path have been determined with respect to the rod coordinate system, which can be converted to the world coordinate system. 

\comment{In practice, our system discards the first and the last sampled points ($\theta=0$ and $2\pi$) and takes samples along the spiral curve to generate the path of the wrapping motion.} The wrapping motion is connected with auxiliary motions to prepare the rope for wrapping and to release it after wrapping, as detailed in \Cref{section:aux}. Fig.~\ref{fig:rviz} illustrates the connection of the wrapping motion to the releasing motion.

\begin{figure}[t]
     \centering
     \subfloat[The gripper rotates $90^\circ$ and moves to the front of the rope.\\ \label{fig:rope_front}]
         {\includegraphics[width=0.45\linewidth]{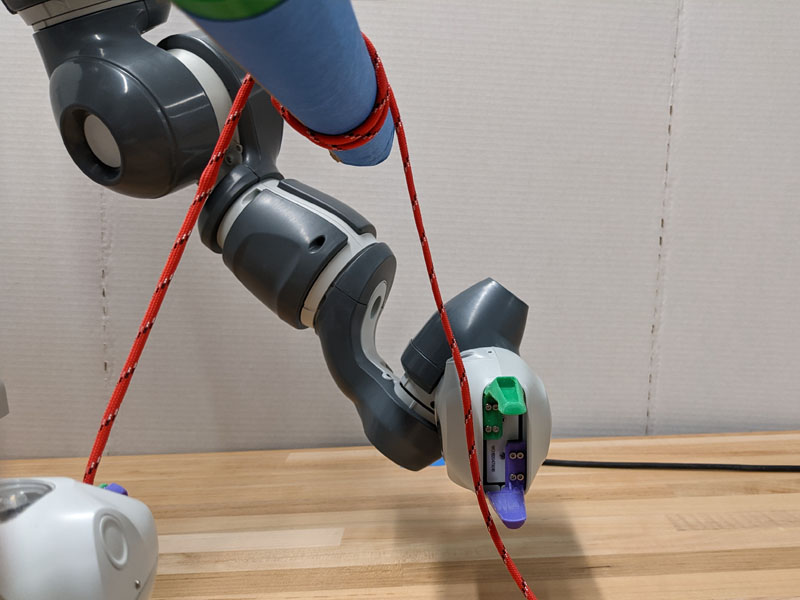} }
     \hfill
     \subfloat[The gripper pushes the active section to align it vertically beneath the rod. \label{fig:push_back}]
         {\includegraphics[width=0.45\linewidth]{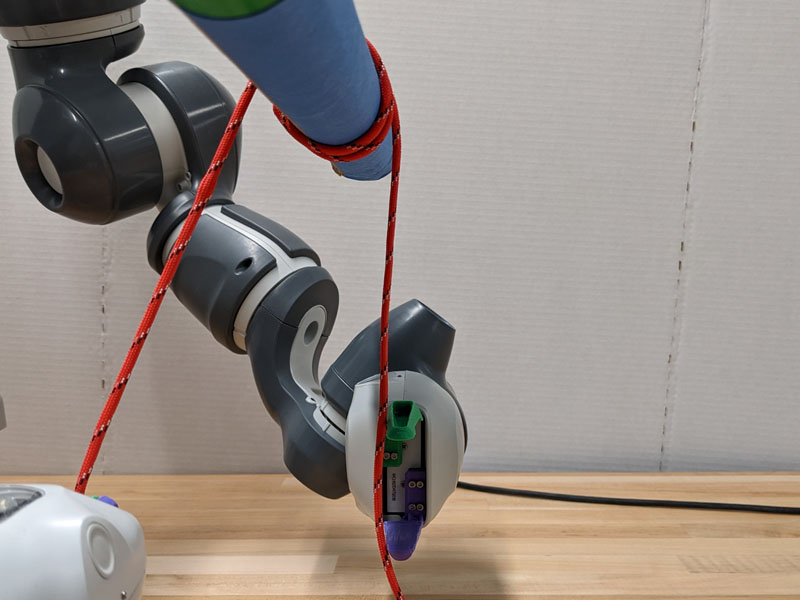} }
     \caption{The auxiliary motions.}
    \label{fig:auxiliary_motion}
\end{figure}

\subsection{Generation of auxiliary motions to facilitate wrapping}\label{section:aux}

Rope picking and releasing motions are defined based on the spiral wrapping motion. Picking is a point-to-point motion that contains three key poses: 1) the entry pose, where the gripper moves to the point with the opening facing toward the rope at a distance. 2) the grasp pose, where the rope is squeezed in between the fingers of the gripper, and 3) the connection pose, where the gripper moves to the starting point of the spiral path. The grasp pose is derived as in \Cref{section:gp_selection} with $L$ being the input value for $l_{gp}$. The entry pose is obtained by offsetting the grasp pose to move the gripper away from the rope.

The release motion is designed to straighten the rope and to move the gripper away from the rope to avoid occlusion for image processing after the wrapping motion is done. Straightening the rope happens after the gripper arrives at the last sampled point of the spiral (see Fig.~\ref{fig:rope_flip} and \ref{fig:rviz}). The gripper \comment{keeps its current orientation and }moves downward to follow a straight-line path, sliding along the rope to flip the active section to the front. Then the gripper releases the rope and withdraws to be away from the rope. At this point, if the remaining length of the active section is longer than the rod's height, part of the section may land on the table, preventing the rope drops vertically from the rod. An additional rope alignment step is added. The left gripper rotates $90^{\circ}$ to increase the contact area with the rope. Then it moves to the front of the active section, as shown in Fig. \ref{fig:rope_front} and pushes the rope towards the manipulator to ensure that the rope is roughly vertical beneath the rod (Fig. \ref{fig:push_back}). Finally, the gripper moves away from the rope to leave space for the next grasp point selection step.

\subsection{\comment{Real-time Perception} and feedback control}\label{section:fb_control}

Following the completion of one wrap, our system takes an image of the result and checks for the tightness along the radial direction (refer to as the height) and the axial direction (refer to as the advance) of the wrap, which is then used for feedback control to improve the next wrap. 

\begin{figure}
     \centering
     \subfloat[The rod and the below area are extracted to examine the height of the last wrap. \label{fig:len_rgb}]
         {\includegraphics[width=0.45\linewidth]{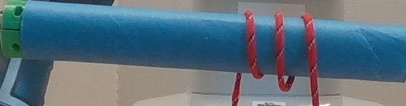} }
     \hfill
     \subfloat[The last wrap is extracted and skeletonized to find the valley point (green). The bottom edge of the rod is indicated by the red line. \label{fig:wrap_valley}]
         {\includegraphics[width=0.45\linewidth]{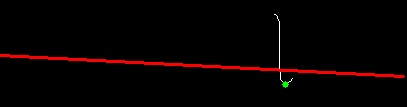} }

    \subfloat[The rod area is extracted to examine the advance of the last wrap. \label{fig:adv_rgb}]
         {\includegraphics[width=0.45\linewidth]{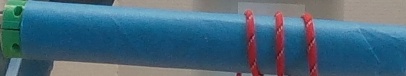} }
     \hfill
     \subfloat[The rope area created by the last wrap (blue) and the gap area (red) are extracted and measured. The rest of the rope area is indicated in white. \label{fig:adv_cluster}]
         {\includegraphics[width=0.45\linewidth]{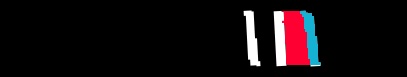} }
         
     \caption{Image processing to examine the \comment{quality} of the last wrap.}
\end{figure}

It starts by checking the height of the last wrapped part of the rope. At this step, a rectangle that contains the rod and some areas below is extracted from the image to include pieces of rope that could possibly hang under the rod, as shown in Fig. \ref{fig:len_rgb}. The rope is extracted from the image by using the rope's hue range as the threshold. Assuming that the rope segments in this binary image share the same width $d$ as the detected rope's diameter, \comment{our system searches the image from right to left, row by row. For each row, it removes the first $d$ white pixels and keeps the next set of connected white pixels with a total number less than $d$. The remaining white pixels belong to the rope segment created by the last wrap.} The segment is skeletonized to find the valley point along it, as shown in Fig. \ref{fig:wrap_valley}. The pixel distance from this point to the bottom of the rod is taken as the height feedback of the last wrap, noted as $h$. This feedback is used to estimate the radius $R$. The feedback controller updates the $R$ by:
\begin{equation}
R_{n+1}=R_n-{\comment{K_{PR}\cdot q_r}}\text{, where }q_r=h-t_R
\end{equation}
\noindent where $K_{PR}$ is the proportional coefficient, $t_R$ is a threshold. The stop condition is when $h\leq t_R$.

The RGB image is also used to check the advance. Pixels of the rod's area (Fig.~\ref{fig:adv_rgb}) are selected and processed to extract the rope. \comment{Like height-checking, the system uses the threshold to create a binary image and scans each row to find the first set of connected white pixels. If the number of pixels is greater than $1.5d$, the system assumes that the last two wraps are in close contact. Otherwise, the set of connected black pixels adjacent to the left of the white pixel set is considered the gap area.} The rope area created by the last wrap $S_r$ and the gap between the last two wraps $S_g$ are measured (Fig.~\ref{fig:adv_cluster}). The feedback controller updates $a$ by:
\begin{equation}
\begin{aligned}
a_{n+1} = a_n-{\comment{K_{Pa}\cdot q_a}}\text{, where }q_a = S_g/(S_g+S_r)
\end{aligned}
\end{equation}
\noindent where $K_{Pa}$ is the proportional coefficient. There are two stop conditions for learning the advance: 1) $q_a=0$, 2) for two consecutive wraps $n$ and $n+1$, $|q_{an}-q_{a(n+1)}|<t_a$, where $t_a$ is a threshold.  When the wrapping output meets either condition, the system stops updating the advance.

\section{Experiments and Results}\label{section:exp_n_result}

In this section, we describe our experiments to implement and test our approach described in \Cref{section:approach,section:perception,section:synergy} and present the results. We then discuss the performance and potential improvement of the wrapping algorithm.

\subsection{Expriment setup}

\setcounter{table}{2}
\begin{table*}[!b]
\caption{Warpping \comment{quality} of each case. The numbers above the wraps indicate the trial number. $a_i$ is the corresponding advance that is used to achieve that wrap.}
\label{tab:exp_result}
\setlength{\tabcolsep}{3pt}
\begin{center}
\begin{tabular}{ | c | c  c  c  | c  c  c | c  c  c | c |}
    \hline
      & \multicolumn{3}{c|}{Rope1} & \multicolumn{3}{c|}{Rope2} & \multicolumn{3}{c|}{Rope3} & Human\\
      & \multicolumn{3}{c|}{     } & \multicolumn{3}{c|}{     } & \multicolumn{3}{c|}{     } & wrapping\\
    \hline
    Rod1 & \includegraphics[width=0.08\textwidth]{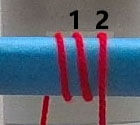}
         & \includegraphics[width=0.08\textwidth]{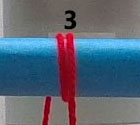}
         &
         & \includegraphics[width=0.08\textwidth]{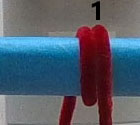}
         &
         &
         & \includegraphics[width=0.08\textwidth]{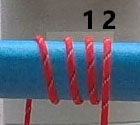}
         & \includegraphics[width=0.08\textwidth]{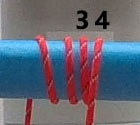} 
         & \includegraphics[width=0.08\textwidth]{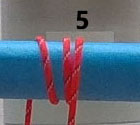} 
         & \includegraphics[width=0.08\textwidth]{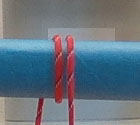}\\
         & $a_1=20.0$ & $a_3=-2.7$ & 
         & $a_1=20.0$ & &
         & $a_1=20.0$ & $a_3=-11.8$ & $a_5=-21.0$ & \multirow{5}{19mm}{Wrapping performed by hands as a comparison.}\\
         & $a_2=11.2$ & (complete) &
         & (complete) & &
         & $a_2=~2.4$ & $a_4=-18.1$ & (complete) & \\
    \cline{1-10}
    Rod2 & \includegraphics[width=0.08\textwidth]{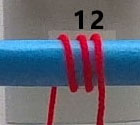}
         & \includegraphics[width=0.08\textwidth]{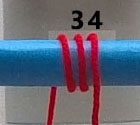}
         & \includegraphics[width=0.08\textwidth]{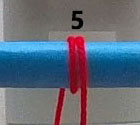}
         & \includegraphics[width=0.08\textwidth]{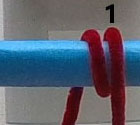}
         & \includegraphics[width=0.08\textwidth]{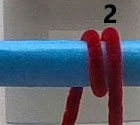}
         & \includegraphics[width=0.08\textwidth]{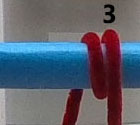} 
         & \includegraphics[width=0.08\textwidth]{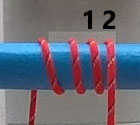}
         & \includegraphics[width=0.08\textwidth]{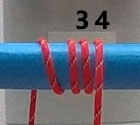}
         & \includegraphics[width=0.08\textwidth]{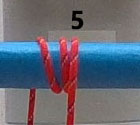} & \\
         & $a_1=20.0$ & $a_3=9.7$ & $a_5=4.7$
         & $a_1=20.0$ & $a_2=17.3$ & $a_3=17.0$
         & $a_1=20.0$ & $a_3=-7.3$ & $a_5=-11.2$ & \\
         & $a_2=10.3$ & $a_4=6.0$ & (complete)
         &  &  & (complete)
         & $a_2=~0.1$ & \comment{$a_4=-7.4$} & (complete) & \\
    \hline
\end{tabular}
\end{center}
\end{table*}
\setcounter{table}{0}

The PC of our system is equipped with an Intel Core i5-7500, 16GB RAM, and an NVIDIA GeForce GTX 1050~Ti, running Ubuntu 20.04 with ROS Noetic. The YuMi was configured to run in manual mode, with 100\% speed, controlled by a modified KTH's YuMi package \cite{kth-ros-pkg}. To achieve the required gripper motion during the wrapping as mentioned in \Cref{section:wrapping}, Axis 6 of the left arm was set to $216^\circ$ at the beginning of wrapping and $-144^\circ$ when finished. Trac-IK \cite{beeson_trac-ik_2015} was used to solve the IK for the remaining 6 axes of the left arm and the 7 axes of the right arm for expected gripper poses. The solver's solving timeout and the error toleration were set to $50$ms and $1$mm, respectively. 

\comment{The system parameters were set as follows. For the fixed section grasp point selection, $l_{gp}=13$mm. The safe distance was $L'\in [20,60]$mm. The system's feedback control coefficients were given as $K_{PR}=0.001$, and $K_{Pa}=0.04$.} The thresholds were set to $t_R=1.5d$ and $t_a=5\%$. The system's advance feedback started with the initial $a_0=20$mm. For a rod with an estimated radius $r_{rod}$, the initial $R_0=1.5r_{rod}$. In practice, with a larger $R$, some waypoints of the spiral path may lie outside the manipulator's workspace, which prevents the manipulator from executing the path. If this happens, $R$ is reduced by a small value, which is set to $5$mm in the experiments, to generate a new spiral path. This process repeats until the manipulator is able to follow the path to get height feedback for the first time. Then the system uses feedback control to tune $R$.

To validate the capability of handling objects with unknown attributes, we tested our system on two cylindrical rods and three ropes, namely Rod1, Rod2, Rope1, Rope2, and Rope3. Two rods share the same length of $l=280$mm but have different radii. Table~\ref{tab:rod_estimation_result} shows the ground truth, the mean and the standard deviation (SD) of the estimated radius for each rod. We ran the estimation 10 times per rod. Rope1 is \comment{complaint and light} yarn (Softee Chunky Solid Yarn, Bernat). Rope2 is \comment{complaint and heavier} yarn (Chenille Home Yarn, Loops \& Threads). Rope3 is \comment{stiff} paracord (1/8 in. x 50 ft. Assorted Color Paracord, Everbilt). We tested our system on the 6 combinations of the rods and ropes. 

\comment{Note that the active section length that the auxiliary motion can handle is limited and varies depending on the rope length. If the remaining rope is not sufficient for more wraps. We enable the manipulator to continue practicing wrapping by unraveling the rope manually and letting the manipulator repeat the whole process.}

\begin{table}[tb]
\caption{The ground truth and the estimation of $r_{rod}$ (over 10 times)}
\label{tab:rod_estimation_result}
\begin{center}
\begin{tabular}{ | l | c | c | c |}
    \hline
     & Ground truth& Estimation mean(SD) &  Range \\
    \hline
    Rod1 & $21$mm & $18.5(\pm1.8)$mm & $17.4\sim 20.7$mm\\
    \hline
    Rod2 & $17$mm & $14.3(\pm1.1)$mm & $13.6\sim 14.6$mm \\
    \hline
\end{tabular}
\end{center}
\end{table}

\begin{table}[tbp]
\caption{The system selected parameter values for radial tightness}
\label{tab:r_tightness}
\begin{center}
\begin{tabular}{ | c | c | c | c |}
    \hline
     & Estimated radius  $r_{rod}$  &  $R$ & $L'$ \\
    \hline
    Rod1 & $20.7$mm & $21.0$mm & $60.0$mm\\
    \hline
    Rod2 & $14.6$mm & $16.9$mm & $60.0$mm \\
    \hline
\end{tabular}
\end{center}
\end{table}

\subsection{\comment{Wrap quality}}\label{section:wrapping_quality}

We used one estimated radius $r_{rod}$ for each rod to examine the wrapping algorithm. The system uses the method mentioned in \Cref{section:wrapping} to obtain $R$ and $L'$. We find that the system-selected values of the two parameters are rope-independent. The values are shown in Table~\ref{tab:r_tightness}.

\begin{figure}[tb]
    \captionsetup[subfloat]{labelformat=empty}
     \centering
     \subfloat[]
         {\includegraphics[width=0.30\linewidth]{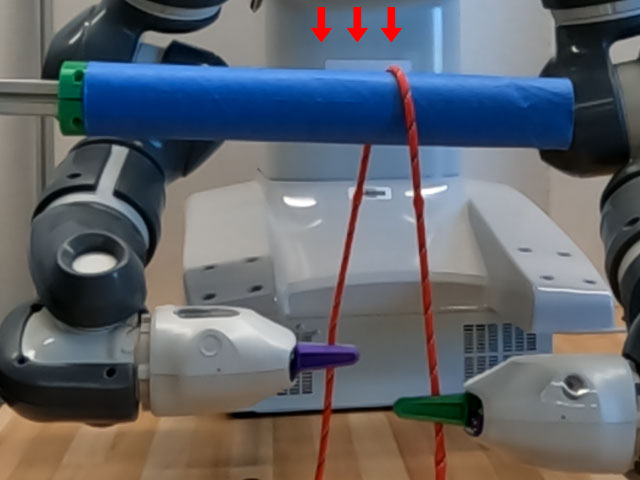}}
     \hfill
     \subfloat[]
         {\includegraphics[width=0.3\linewidth]{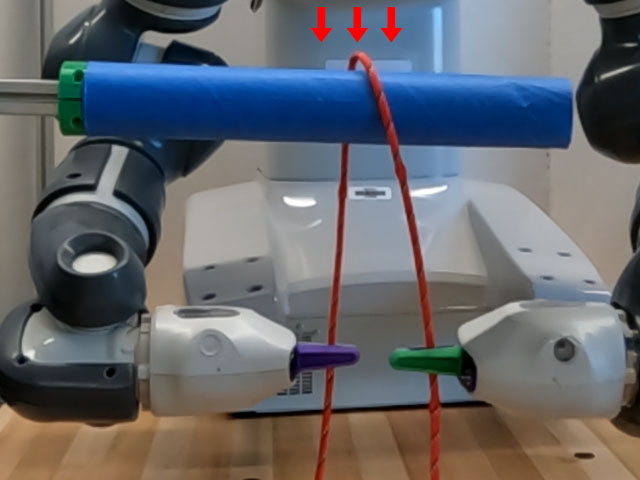} }
     \hfill
    \subfloat[]
         {\includegraphics[width=0.3\linewidth]{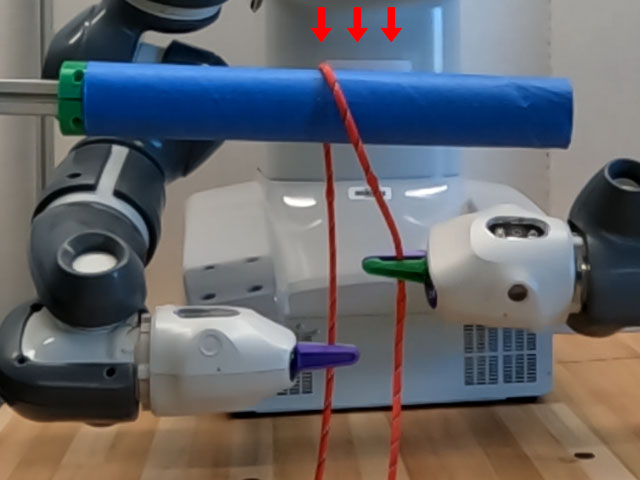} }

     \vspace{-1\baselineskip}
     \caption{The fixed end of the Rope3 is shifted during creating the first wrap because of its stiffness. The red arrows are used as references.}
     \label{fig:first_wrap_pushing}
\end{figure}

The robot took 5 trials to reach the stop condition for learning the advance in the axial direction for \{Rod1, Rope3\}, \{Rod2, Rope1\}, and \{Rod2, Rope3\}. For \{Rod1, Rope1\} and \{Rod2, Rope2\}, it took 3 trials. It achieved the axial tightness in the first trial for \{Rod1, Rope2\}. The result can be found in Table~\ref{tab:exp_result}. These images were taken by the RGB-D camera and were also used for the feedback control process. Note that since Rope3 has a higher stiffness, making the first wrap tends to push the fixed section to create a larger gap compared to the following wraps (see Fig.~\ref{fig:first_wrap_pushing}). This only happens to Rope3 when no previous wrap is made. Therefore, we always kept one pre-wrap on the rod that was not taken into account when testing Rope3 on both rods.

For all cases, the tightness along the radial direction was met. For cases with Rope1 and Rope2, the axial tightness was also met. For cases with Rope3, the final result was comparable with the wrapping result by a person manually (see the last column of Table~\ref{tab:exp_result}).

\comment{The attached video first demonstrates the feedback control process described in \Cref{section:fb_control}, referred to as the training session. It next shows how the system performs after it is trained with the adjusted parameter values to verify the stability, referred to as the testing session.} 

\subsection{Time efficiency analysis} \label{section:time_consumption}

We report the time consumption of our method as the computation time and the total execution time. The former includes all the calculation steps that are described in both \Cref{section:perception,section:synergy}. The latter includes the computation time for \Cref{section:synergy} and execution of the picking, wrapping, and releasing motions. All measurements are repeated 10 times.

Our system spends $13.197(\pm 0.402)$s for the rod estimation and $1.042(\pm 0.053)$s for the rope estimation at the beginning of a wrapping task. These are the pre-processing steps before wrapping and only need to be done once for each combination. \comment{The average time spent for the major procedures of creating one wrap is as follows: grasp point selection takes $4.447(\pm0.082)$s, wrapping motion adjustment and auxiliary motion generation take $2.302(\pm0.050)$s, real-time perception and feedback control takes $0.312(\pm0.016)$s.}

We measured the total execution time of one wrap for \{Rod1, Rope1\}. The mean is $63.933(\pm 0.611)$s, which is an order of magnitude higher than the computation time for motion planning. \comment{Therefore, computation is much faster than the physical wrapping motion.} 

\subsection{Discussion}

We have observed that the radial feedback meets the stop condition from the first wrap for all test cases. Due to the camera’s perspective and the methods mentioned in \Cref{section:rod_estimation}, the estimated $r_{rod}$ is always smaller than the ground truth. With $R$ being less or equal to the ground truth, the gripper slides slightly down the active section of the rope during the wrapping to compensate for the insufficient rope length from the gripper to the rod. This motion keeps the tension of the section and results in radial tightness. This suggests that to wrap over a solid of revolution, an $R$ that is smaller than the ground truth can be tolerated and often helpful for radial tightness.

\section{Conclusions}\label{section:conclusions}
In this paper, we present a novel method \comment{of perception-motion synergy} to solve the problem of using a general-purpose robot manipulator with a parallel gripper to wrap ropes around a rigid rod. This method is based on a parameterized canonical motion and does not require prior knowledge of the rope or the rod. It uses RGB-D images to estimate the state of the rope and rod, evaluates the \comment{quailty} of each wrap, and generates feedback from the evaluation to improve motion planning. We tested our method with 6 combinations of ropes and rods. The result shows that our general method applies to different ropes and rods very well.
\comment{However, our current method estimates wrapping quality based on colors and cannot handle two wraps that cross each other. In addition, the perception step only happens before and after a wrapping motion; thus, information during the wrapping motion is not perceived, which could be informative to wrapping motion adjustment.}
In the next step, we will expand our method to using other types of rods, for instance, solids of revolution other than cylinders, such as a cone, and to incorporate \comment{force control} and tactile sensing \comment{to achieve more accurate and robust perception. We will also improve the efficiency of motion planning}.

\addtolength{\textheight}{-12cm}   






\bibliographystyle{IEEEtran}
\bibliography{IEEEabrv,references}

\end{document}